\documentclass[conference]{IEEEtran}
\IEEEoverridecommandlockouts
\usepackage{cite}
\usepackage{amsmath,amssymb,amsfonts}
\usepackage{algorithmic}
\usepackage{graphicx}
\usepackage{textcomp}
\usepackage{xcolor}
\usepackage[numbers]{natbib}

\usepackage{times}
\usepackage{epsfig}
\usepackage{float}
\usepackage[symbol]{footmisc}
\usepackage{subcaption}
\usepackage{wrapfig}

\def\BibTeX{{\rm B\kern-.05em{\sc i\kern-.025em b}\kern-.08em
    T\kern-.1667em\lower.7ex\hbox{E}\kern-.125emX}}
\begin{document}

\title{Robustness of Bayesian Neural Networks to White-Box Adversarial Attacks
}

\author{\IEEEauthorblockN{Adaku Uchendu*\thanks{*Work done while this author was an intern at Air Force Research Lab (through ATRC 2019)}
}
\IEEEauthorblockA{\textit{College of Information Sciences and Technology} \\
\textit{Pennsylvania State University}\\
University Park, PA, USA \\
azu5030@psu.edu}
\\
\IEEEauthorblockN{Christopher Menart}
\IEEEauthorblockA{\textit{Sensors Directorate} \\
\textit{Air Force Research Lab} \\
Dayton, OH, USA \\
christopher.menart@us.af.mil}
\and
\IEEEauthorblockN{Daniel Campoy}
\IEEEauthorblockA{\textit{Applied Research Solutions, Beavercreek, OH} \\
\textit{Air Force Research Lab}\\
Dayton, OH, USA \\
dcampoy@appliedres.com}
\\
\IEEEauthorblockN{Alexandra Hildenbrandt} 
\IEEEauthorblockA{\textit{Sensors Directorate} \\
\textit{Air Force Research Lab} \\
Dayton, OH, USA \\
alexandra.hildenbrandt.1@us.af.mil}
}

\maketitle

\begin{abstract}

Bayesian Neural Networks (BNNs), unlike Traditional Neural Networks (TNNs) 
are robust and adept at handling adversarial attacks by incorporating 
randomness. This randomness improves the estimation of uncertainty, a feature lacking in TNNs.  
Thus, we investigate the robustness of BNNs to white-box attacks using multiple Bayesian neural architectures.
Furthermore, we create our BNN model, called \textit{BNN-DenseNet}, by fusing Bayesian inference (i.e., variational Bayes) to the DenseNet architecture, and BDAV, by combining this intervention with adversarial training. 
Experiments are conducted on the CIFAR-10 and 
FGVC-Aircraft datasets. 
We attack our models with strong white-box attacks ($l_\infty$-FGSM, $l_\infty$-PGD, $l_2$-PGD, EOT $l_\infty$-FGSM, and EOT $l_\infty$-PGD). 
In all experiments, at least one 
BNN outperforms traditional neural networks during 
adversarial attack scenarios. An adversarially-trained BNN outperforms its non-Bayesian, adversarially-trained counterpart in most experiments, and often by significant margins. 
These experimental results suggest that the dense nature of DenseNet provides robustness advantages 
that are further amplified by fusing Bayesian Inference with the architecture. 
Lastly, we investigate network calibration and find that BNNs do not make overconfident predictions, providing evidence that BNNs are also better at measuring uncertainty.
\end{abstract}


\section{Introduction}

Deep learning has successfully been applied to several fields and  tasks. It  has  achieved  the  most  notoriety  on  image classification and object detection tasks. However, recently, it has been applied to more sensitive fields such as the Medical field,  where  mis-classification  can  be  catastrophic.  This does not seem to be a problem  since  deep  learning  achieves  state-of-the-art performance  in the many  
tasks  and  fields  it  has  been applied to.
Surprisingly, while being powerful, it is  
vulnerable to adversarial attacks \cite{szegedy2013intriguing, goodfellow2014explaining, fawzi2018adversarial}. 
In fact small perturbations or changes to data, which are often 
imperceptible to the human eye is enough to cause a high-performing 
deep learning model to misclassify \cite{fawzi2018adversarial}.

Therefore, we hypothesize that certain aspects of the
deep learning framework make it vulnerable to 
adversarial attacks. 
Specifically, the use of single-point estimates cause 
traditional Neural Networks (NNs) to be susceptible 
to adversarial attacks. 
\citeauthor{nguyen2015deep} and \citeauthor{shuklaadversarial} 
claim that these single-point estimates limit the model's
decision boundary, 
which leads to decisions outside the boundary, making the NN vulnerable to attacks.
This suggest that
NNs are not well-calibrated since, they overfit and make overconfident 
predictions \cite{hernandez2015probabilistic}, which 
allows the model to perform very well while still being 
vulnerable to 
adversarial attacks. To mitigate these issues, we propose
fusing Bayesian Inference (i.e., variational Bayes) with traditional NNs. We claim that this Bayesian 
approach allows NNs to estimate for uncertainty better, which leads to
production of less overconfident predictions.

Our ultimate goal is to build a defense model (Bayesian 
Neural Network (BNN)) robust to white-box 
adversarial attacks. 
In the White-box adversarial attack scenario, the attacker has knowledge of the model parameters. 
Naturally, to further improve the robustness of our BNN, 
we propose a novel methodology,  \textit{BDAV} (BNN-DenseNet + Adversarial training), 
which incorporates adversarial training \cite{madry2017towards} with BNN. 
\textit{BDAV} bears a similarity to Adv-BNN (adversarially trained BayesianVGG) \cite{liu2018adv} but differs in the backbone, training and parameter choices. 
In our implementation of \textit{BDAV}, we show
that this defense model, unlike others, decays linearly
at a slow rate when attacked with strong adversarial perturbations.
This is possibly due the dense nature of DenseNet. We observe that fusing 
Bayesian Inference with DenseNet amplifies the advantages of 
a dense neural network. Furthermore, incorporating adversarial training on 
this BNN, improves the model's robustness as well. 
\textit{BDAV} achieves state-of-the-art results 
when attacked with strong white-box attacks.

We test our hypothesis by training 6 models (DenseNet, VGG, BayesianVGG, BNN-DenseNet, BDAV, and Adv-DenseNet) with 2 datasets - CIFAR-10 and FGVC-Aircraft \cite{maji13fine-grained}. Then, we attack the trained models to test robustness with 5 state-of-the-art white-box attacks
($l_\infty$-FGSM, $l_\infty$-PGD, $l_2$-PGD, EOT $l_\infty$-FGSM, and EOT $l_\infty$-PGD). Expectation Over Transformation (EOT)
creates different transformations of an image such that the slightest of shift can fool the classifier \cite{athalye2018synthesizing}. Lastly, our key contributions can be summarized as:
\begin{itemize}
    \item We show that fusing Bayesian Inference and traditional NN framework  (DenseNet121) improves the robustness of the traditional NN.
    In fact, in order to activate all/most of the benefits of Bayesian Inference, it has to be fused with the right traditional NN. \citeauthor{liu2018adv} used VGG16, however, we find that the dense nature of DenseNet121 improves the results.  
    
    \item We observe that optimizer choices, such as using Adam optimizer, positively affect the model. This suggest that the natural accuracy of a BNN is a robustness measure.
    

    \item We show experimental results on a large dataset, FGVC that has not been used for this kind of task (i.e., adversarial robustness).
    
    \item We show that BNNs are even robust to EOT attacks, an attack robust to stochastic models. 
    
    \item Lastly, we perform a network calibration analysis on all the models and we plot the accuracy's output and confidence (probabilities of each class) into an identity function. Identifying how confident each model is in its prediction of clean data.
    This provides further insight on which models need more or less calibration.
\end{itemize}


\begin{figure*}[htb] 
    \centering
    \begin{subfigure}[b]{\textwidth}
        \centering
        \includegraphics[width=0.29\linewidth]{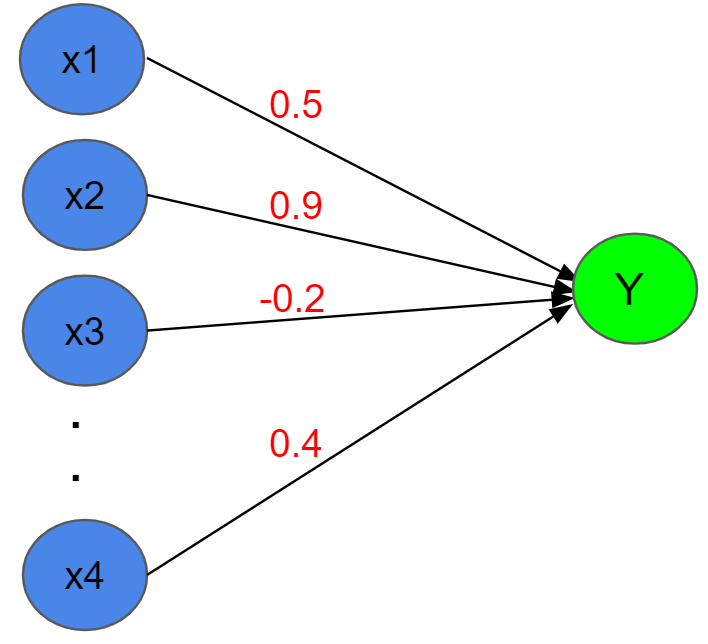}
        \includegraphics[width=0.28\linewidth]{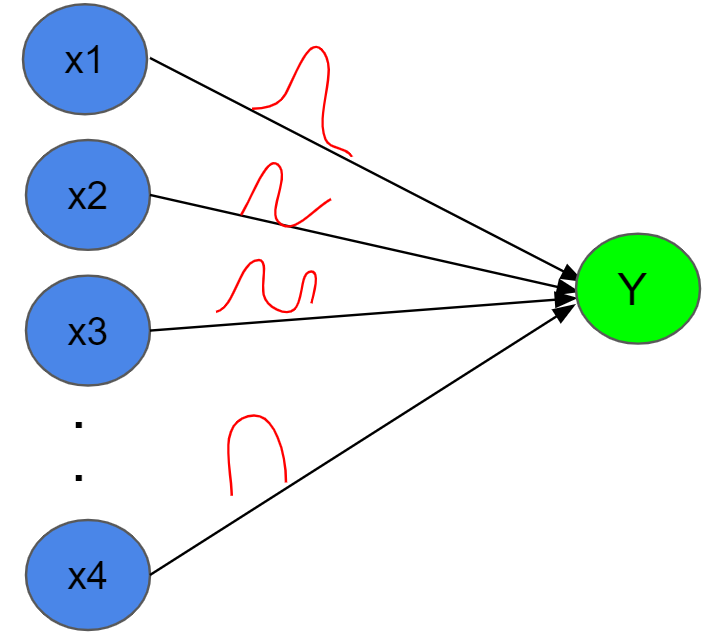}
    \end{subfigure}

    \caption{Traditional Neural Network (left) vs. Bayesian Neural Network (right).}
    \label{fig:tnn}
\end{figure*}

\section{Related Work}
In the quest to improve the adversarial robustness of neural networks, a few researchers adopt fusing Bayesian Inference with 
traditional NNs. 
\citet{ye2018bayesian} proposes \textit{BAL} (Bayesian Adversarial Learning) to alleviate the shortcomings of traditional NNs. 
The posterior of the BNN is approximated with a scalable Markov Chain Monte Carlo sampling strategy, specifically 
stochastic gradient adaptive Hamiltonian Monte Carlo (SGAdaHMC) \cite{ye2018bayesian}. Thus to construct \textit{BAL}, the BNN is
trained with sampled adversarial examples 
and significantly outperforms all other defense models (i.e., 
adversarially trained traditional NNs). \textit{BAL} is further improved by \textit{Adv-BNN}, 
which is proposed by \cite{liu2018adv}.
The posterior of this BNN is approximated with 
variational Bayes/Inference, specifically using 
a Gaussian distribution \cite{liu2018adv}. 
This BNN approach is trained with
adversarial examples generated by $l_{\infty}$-PGD attack. 
\cite{liu2018adv, ye2018bayesian} show that 
adversarial training of BNNs further improves robustness and achieves 
state-of-the-art results in defending against adversarial attacks. 

Another technique to improve a traditional NN involves
using Kronecker factored Laplace to approximate the posterior of a NN
\cite{ritter2018scalable}. 
When evaluated on white-box adversarial attacks, these BNN perform better at 
measuring uncertainty and are unlikely to make false predictions. 
Furthermore, the results suggest that it is less vulnerable to both 
targeted and non-targeted adversarial attacks 
\cite{ritter2018scalable}. 
Additionally, using 2 
BNNs (i.e., Variational Inference and Hamilton Monte Carlo) trained on 
MNIST and attacked with $l_{\infty}$-FGSM and $l_{\infty}$-PGD attacks, 
BNNs are found to be robust to these attacks \cite{carbone2020robustness}. 
This work focuses on proving that BNNs should be robust to gradient-based attacks. 
Furthermore, 
\cite{carbone2020robustness} theoretically prove that 
gradient-based attacks in the thermodynamic limit are ineffective to 
BNNs. 

Lastly, BNN's ability to measure uncertainty well makes it robust to 
adversarial attacks \cite{cardelli2019statistical, 
smith2018understanding} due to the statistical guarantees leveraged
by it's ability to withstand adversarial attacks.

\section{Bayesian Neural Network} \label{BNN}
A Neural Network (NN) can be represented as a probabilistic model 
$p(y|x,w)$ such that $y$ is the set of classes and 
$p(y|x,w)$ is a categorical distribution. 
Given training dataset $D=\{x^{(i)},y^{(i)}\}$, we can obtain the weights 
$w$ of a NN by maximizing the likelihood function $p(D|w) = \prod_i 
p(y^{(i)}|x^{(i)}, w)$. 
Maximizing the likelihood function gives us the MLE (Maximum 
Likelihood Estimate) of $w$. 
Thus, to transform a traditional Neural Network into a 
Bayesian Neural Network, we use Bayes theorem:
\begin{equation}
    p(w| D) = \frac{p(D|w) p(w)}{p(D)}
\end{equation}
to estimate the weights.
$p(w|D)$ is the probability of the weights given the data. It is known 
as the \textit{posterior} probability. 
$p(D|w)$ is the \textit{likelihood}, $p(w)$ is 
the \textit{prior}, and $p(D)$ is the \textit{evidence}
probability. Using bayes theorem, 
we can get a probability distribution that estimates the 
weights instead of single-point estimates gotten from MLE. 
But solving for 
$p(D)$ expands to a high dimensional integral,
$p(D) = \int p(D, w) dw = \int p(w) p(D|w) dw$
\cite{blei2017variational}. This makes solving for an analytical 
solution of $p(w|D)$ intractable. Thus, an approximate function is 
needed to approximate the true posterior. Some of the techniques include 
Metropolis-Hastings, Hamilton Monte Carlo, Variational Inference, Etc. \cite{baye}. 
We will focus on variational Bayes/inference 
\cite{blundell2015weight, hinton1993keeping, graves2011practical}, 
the technique we employed because of its inexpensive computational 
cost and unbiased estimates it samples. 
Using the variational Bayes technique,  
a variational distribution $q(w|\theta)$ of known functional form is used 
to approximate the true posterior. 
This is achieved by minimizing the Kullblack-Leibler divergence
between $q(w|\theta)$ and the true posterior $p(w|D)$ w.r.t. $\theta$, which is 
represented as $KL(q(w|\theta) ~||~ p(w|D))$. The KL-divergence can be expanded to:
\begin{equation}
    F(D, \theta) = KL(q(w|\theta) ~||~ p(w)) - \mathbb{E}_{q(w|\theta)} 
    \log(p(w|D))
\end{equation}
where,
\begin{equation*}
    KL(q(w|\theta) ~||~ p(w)) = \int q(w|\theta) \log \frac{q(w|\theta)}{p(w)} dw
\end{equation*}
However, this too becomes intractable since finding an analytical solution for
$KL(q(w|\theta) ~||~ p(w))$ is too computationally costly to solve in real-time. 
Therefore, to solve for the posterior, we can sample from the approximate function,
$q(w|\theta)$. This is easier than sampling from the true posterior,
$p(w|D)$. In doing so, we achieve a tractable function that can be written as:
\begin{equation}
    F(D, \theta) \approx \frac{1}{N} \sum_{i=1}^N  \big[ \log q(w^{(i)}|\theta) 
    - \log p(w^{(i)}) - \log p(D|w^{(i)}) \big].
\end{equation}

To sample $q(w|\theta)$, we model $\theta$ as a Gaussian distribution such that 
$\theta \sim \mathcal{N}(\mu, \sigma^2)$,
where $\mu$ and $\sigma^2$ are the mean and variance vectors of the distribution, respectively. 
Approximation of the true posterior using KL-divergence is also known
as the ELBO (Evidence Lower BOund) method since the goal is to 
maximize the evidence lower bound \cite{zhang2018noisy}. See difference between Traditional NN and BNN in figure \ref{fig:tnn}. 

Lastly, training a BNN differs slightly from training a traditional
NN. In traditional NNs, the weights and biases are calculated and updated with 
backpropagation. Since BNN has 2 parameters ($\mu$ and 
$\sigma^2$) to estimate the weights, the training process requires 
two parameters to be calculated and updated. This training process is 
is known as \textit{Bayes by Backprop} \cite{blundell2015weight}. 



\section{Method}
Since the ultimate goal is to build a model robust to adversarial 
attacks, specifically white-box attacks, we investigate the role of 
fusing Bayesian Inference with traditional NN. 
Previous work fused Bayesian Inference with VGG16 \cite{liu2018adv},
and we hypothesize that we could build a more robust BNN by changing the 
backbone and making different parameter choices. 
VGG16 \cite{simonyan2014very}, while a powerful image 
classifier, does not possess some of the advantages of DenseNet  \cite{li2018densely},
which makes it a robust classifier. 
These advantages include the use of densely connected layers to diversify features \cite{zhu2017densenet}, reduction of parameters \cite{zhu2017densenet}, and alleviation of the vanishing-gradient
problem \cite{jegou2017one}.
Therefore, we fuse Bayesian Inference with DenseNet121, creating
\textit{BNN-DenseNet}, making the backbone of our BNN, DenseNet.
We choose DenseNet121 instead of other variants of DenseNet because 
it has fewer dense layers making it suitable for extracting learned concepts.
Fusing Bayesian Inference with DenseNet121 means that 
the Convolutional and Batchnorm layers of DenseNet will no longer 
have single-point estimates, but a Gaussian distribution, $N(0_{d}, 
\sigma^2 I_{d \times d})$, to sample the posterior's weights and biases. 
Therefore, weights are dependent on 2 parameters  
$w \sim \mathcal{N}(\mu, \sigma^2)$
Furthermore, this essentially doubles 
NN parameters. The use of a Gaussian distributions of different sizes to estimate the weights and biases of the NN leads to the training of an ensemble of NNs. 

We observe that the benefits of BNNs: (1) naturally regularizes; 
(2) does not overfit; (3) measures uncertainty well; 
(4) trains an ensemble of NNs. 
Therefore, we decide to further improve upon our
BNN by training it with adversarial examples generated
with $l_\infty$-PGD using $\epsilon=0.03$ and 10 iterations. 
Transforming BNN-DenseNet 
$\to$ BDAV is a two-fold problem - (1) generate adversarial data, 
$D^{adv}$ such that new data can fool classifier, and (2) learn from 
adversarial data, $D^{adv}$ such that predictions are reasonable.
This leads to having a loss function that can capture the two-fold
problem: (1) loss function for predicting the perturbed data and 
(2) generating the adversarial examples. Which can be written as:
\begin{equation}
    L(D) = cross\_entropy + \beta \cdot KL\_loss,
\end{equation}
where $\beta$ is a hyperparameter for balancing the cost of 
data generation. This is also known as the KL-divergence weight ($kl\_weight$).


\begin{figure}
 \centering
    \includegraphics[width=1.1\linewidth]{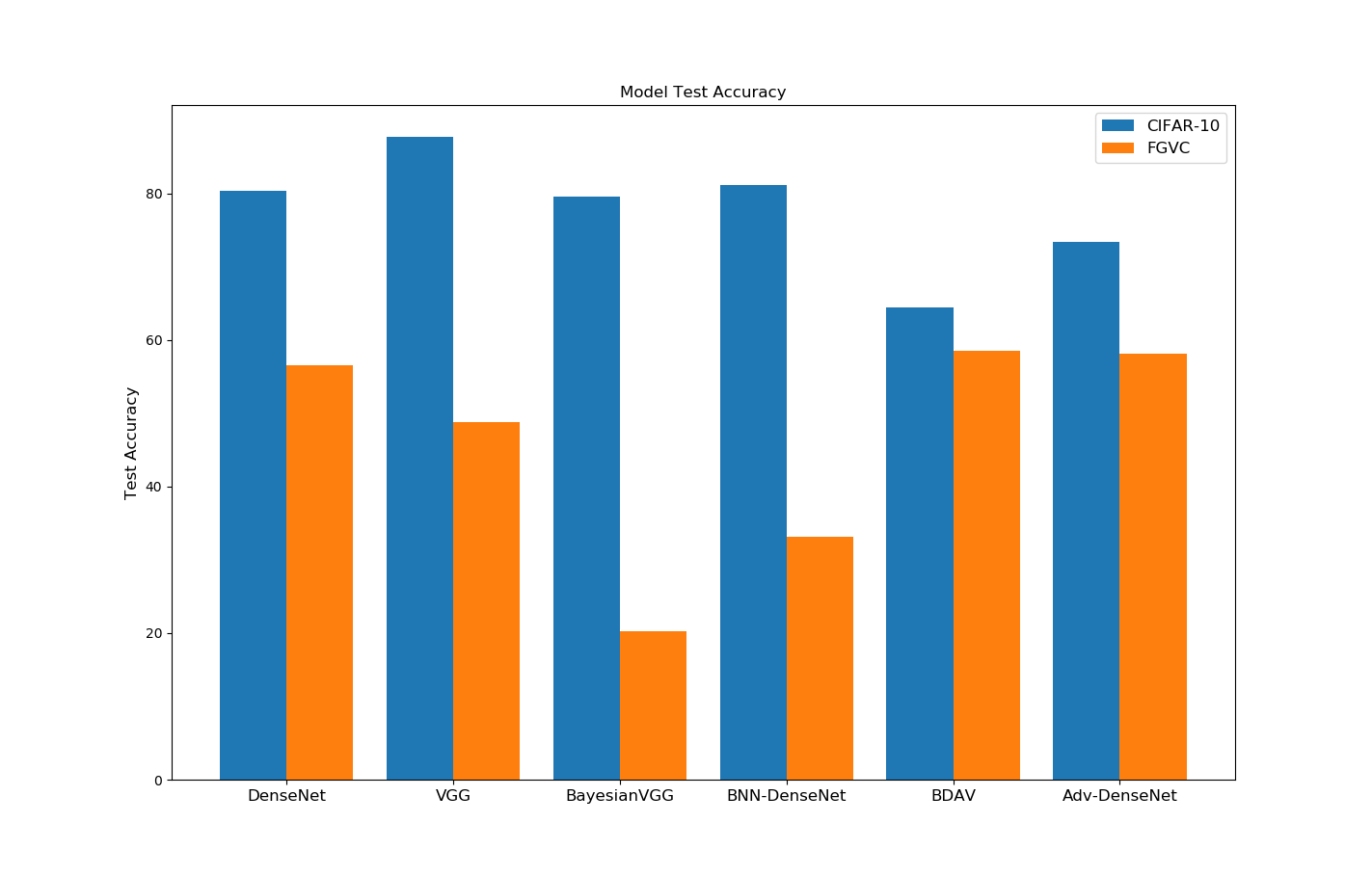}
    \caption{Test Accuracy of the 6 models trained on CIFAR-10 
    and FGVC datasets}
    \label{fig:acc}
\end{figure}

\section{Experimental setup}
\subsection{Data}
For our experiments, we use benchmark datasets - CIFAR-10 \cite{krizhevsky2009learning} and FGVC-Aircraft (Fine-Grained Visualization Classification) \cite{maji13fine-grained} datasets. 
CIFAR-10 and FGVC are 
$32 \times 32$ pixels and $224 \times 224$ pixels, respectively.
Thus, FGVC is the more non-trivial dataset with 10,000 Aircraft images.
Furthermore, its 
labels are hierarchical, and it could be classified with 100 airplane models, 70 families, or 30 manufacturers as labels. For this task, we use the 30 manufacturers (i.e., Boeing, Airbus, Eurofighter, Etc.) labels to classify the dataset. 

\subsection{Model Architecture}
We use Adam optimizer for all the models except BayesianVGG, which used Stochastic Gradient Descent as its optimizer.
\begin{itemize}
    \item \textbf{DenseNet}\footnote[1]{baseline model}: We use DenseNet121 \cite{huang2017densely},
    a variant of DenseNet. The model was implemented using \textit{torchvision}, a python package 
    with the default parameters (i.e., \# of input features are 32).
    
    \item \textbf{VGG}$^*$: We use an off-the-shelf VGG16 \cite{simonyan2014very} model from the \textit{PyTorch} python package. 
    
    \item \textbf{BayesianVGG}$^*$: Obtained from \citeauthor{liu2018adv}'s implementation of fusing variational inference/Bayes with VGG16. The variance and mean were chosen to be $0.15$ each. 
    
    \item \textbf{BNN-DenseNet}: This is our contribution. We use \textit{torchbnn}\footnote[2]{https://github.com/Harry24k/bayesian-neural-network-pytorch}, a python package to fuse Bayesian inference with DenseNet121. Previous experiments on DenseNet suggested robustness \cite{jang2019densenet, li2018densely},
    and thus fusing it with Bayesian inference improved the robustness even further. The variance and mean were chosen to be $0.15$ each. 
    
    \item \textbf{BDAV}: This is also our contribution. BDAV is BNN-DenseNet trained with adversarial examples generated with $l_{\infty}$-PGD attack with 0.03 epsilon and 10 iterations. 
    We use Adam optimizer, constant kl weight (0.1) and default parameters of DenseNet121 to train BDAV. 
    
    \item \textbf{Adv-DenseNet}$^*$: We train on DenseNet121 with adversarial examples generated by $l_{\infty}$-PGD attack with 0.03 epsilon and 10 iterations. 
    This was to compare the robustness of adversarially trained traditional NNs vs. adversarially trained BNNs. 

\end{itemize}


\subsection{White-Box Adversarial Attacks}
We attacked our models with strong white-box attacks ($l_\infty$-FGSM, $l_\infty$-PGD, $l_2$-PGD, EOT $l_\infty$-FGSM, and EOT $l_\infty$-PGD), which are variants of: 
\begin{itemize}
    \item \textbf{FGSM}. This is a gradient-based attack introduced by \cite{goodfellow2014explaining}. Fast Gradient Sign Method (FGSM) works by applying perturbations in the direction of the gradient. This can be written as:
    \begin{equation}
        x^{adv} = x + \epsilon \cdot sign(\nabla_{x} J(x, y_{target})),
    \end{equation}
    where $x$ is the clean image, $x^{adv}$ is the adversarially perturbed image, $J$ is the loss function, $y_{target}$ is the target/true label, and $\epsilon$ is the turnable parameter. 
    
    \item \textbf{PGD}. This is another gradient-based attack introduced by \cite{madry2017towards}. Projected Gradient Descent (PGD) is an iterative variant of FGSM making it a stronger attack. 
    This can be written as:
   \begin{align}
        x^{adv_{t+1}} &=  x^{adv_{t}} + \alpha \cdot sign(\nabla_{x} J(x^{adv_{t}}, y_{target})) \\
        x^{adv_{t+1}} &= \textbf{clip}(x^{adv_{t+1}}, x^{adv_{t}} - \epsilon, x^{adv_{t}} + \epsilon)  
   \end{align}
   where $\textbf{clip}(.,a,b)$ function is in the range $[a,b]$. 
   
    \item \textbf{EOT}.  \citet{athalye2018synthesizing} introduced the Expectation over Transformation (EOT) attack. EOT attack generates different image transformations, such that a slightly 
    tilted image can fool the classifier.
   Unlike other attack methods, EOT keeps the distance between the clean and perturbed images in a threshold rather than minimize it
   \cite{athalye2018synthesizing}. The expected distance under transformation can be written as: 
    \begin{equation}
        \mathbb{E}_{t \sim T}[d(t(x^{adv}), t(x))],
    \end{equation}
    where $t(x)$ is the transformed clean image and $t(x^{adv})$ is the transformed adversarial image. The EOT optimizes the search 
    for adversarial examples by maximizing the probability for the true class $y_{target}$ across all possible distributions of transformations T \cite{molnar2020interpretable}. This is written as:
    \begin{equation}
       \arg\max_{x^{adv}} \mathbb{E}_{t \sim T} [\log p(y_{target}|t(x^{adv}))] 
    \end{equation}
    such that the expected distance over all possible transformations
    between $x$ and $x^{adv}$ is constrained at $\epsilon$, the threshold. This can be written as:
    \begin{equation}
       \mathbb{E}_{t \sim T}[d(t(x^{adv}), t(x))] < \epsilon~~
       and~~x \in [0,1]^{d}.
    \end{equation}
    Thus, we investigate the performance of combining the EOT attack with the $l_{\infty}$ attacks. 
    This implies that EOT transformed images perturb either $l_{\infty}$-FGSM or $l_{\infty}$-PGD attacks to improve the attacks' strength further.  
   Lastly, since the EOT attack is robust to models that incorporate randomness, we investigate 
   our BNN models' robustness to this attack.
\end{itemize}

All the $l_{\infty}$ attacks use epsilons in range 
$0-0.07$, while the $l_2$-PGD use epsilons $0-4$. We choose these epsilons 
because they increase the strength of the attacks. 
Thus, we attack our models with 10 and 40 iterations of the $l_\infty$-PGD and $l_2$-PGD attacks. We also test the model performance when we attack with 10-100 iterations of the $l_\infty$-PGD attack with $0.03$ epsilon. 
Lastly, we use 30 ensemble size and 40 iterations for the EOT attacks.

\subsection{Network Calibration analysis}
Recent/modern NNs have achieved higher accuracies than NNs from
past decades. However, while NNs have improved in classification 
and detection tasks, they have also become 
less well-calibrated  
\cite{guo2017calibration}. 
A well-calibrated NN produces confidence output that matches the accuracy. 
For instance, if a NN achieves a 60\% accuracy (i.e., 60\% correct
predictions), a well-calibrated model will also have achieved a 
60\% confidence score for each prediction. 
Thus, to further investigate our trained models' robustness, we plot the accuracy's output and confidence (probabilities of each class) into an identity function. Identifying how confident each model is in its prediction of clean data. 
These plots are also known as 
\textit{reliability diagrams} \cite{niculescu2005predicting, guo2017calibration}.
Furthermore, we use common calibration metrics - ECE (expected calibration error) 
and MCE (maximum calibration
error) to evaluate these plots \cite{guo2017calibration}. 
We adopt \citeauthor{guo2017calibration}'s formal definitions of Accuracy, Confidence, ECE and 
MCE for this analysis. 
Lastly, we did not apply any calibration techniques to our model 
for this work since the aim is to observe which models require more or less calibration. 
Thus, we include \textit{confidence calibration} measures as a robustness metric.

\begin{figure*}[htb!] 
    \centering
    \begin{subfigure}[b]{\textwidth}
        \centering
        \includegraphics[width=0.329\linewidth]{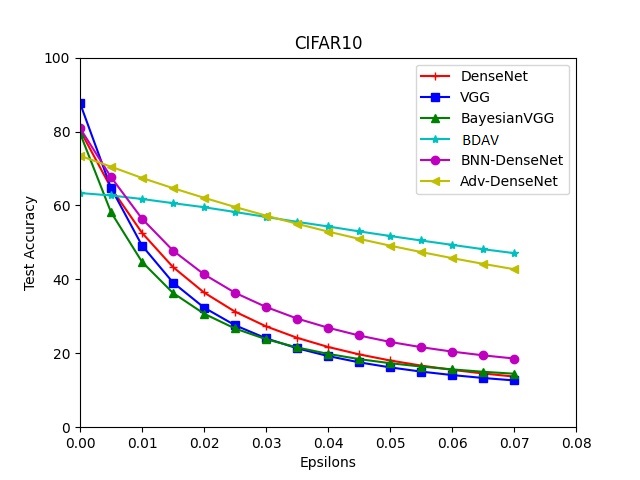}
        \includegraphics[width=0.329\linewidth]{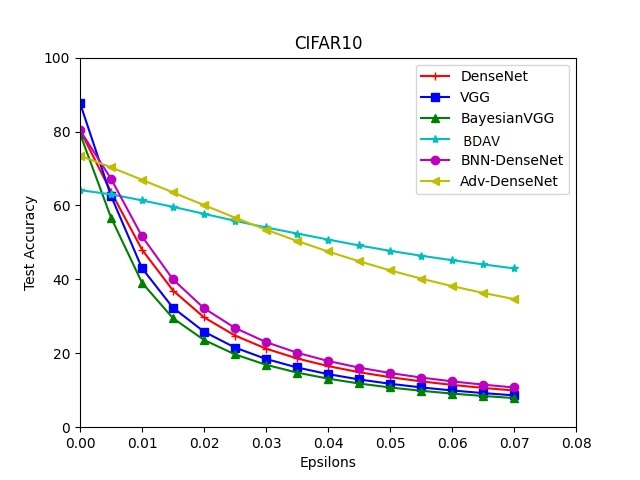}
        \includegraphics[width=0.329\linewidth]{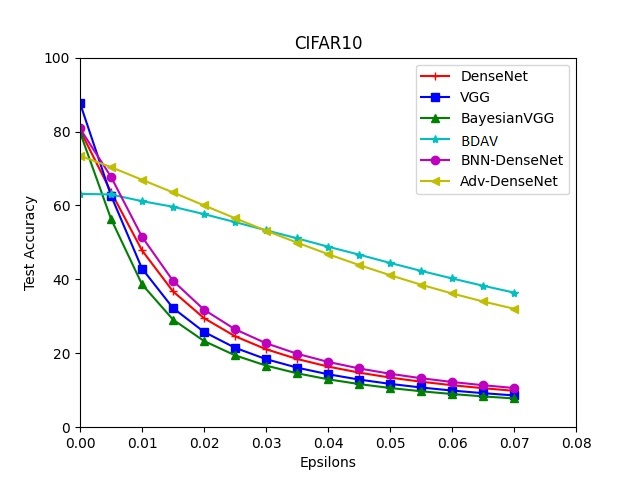}
    \end{subfigure}
    \begin{subfigure}[b]{\textwidth}
        \centering
        \includegraphics[width=0.329\linewidth]{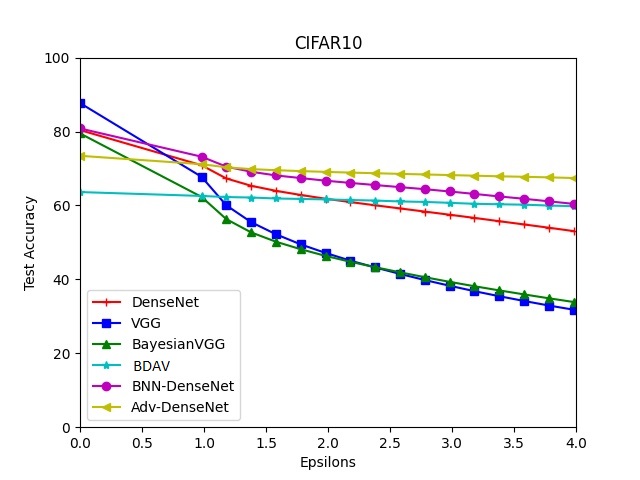}
        \includegraphics[width=0.329\linewidth]{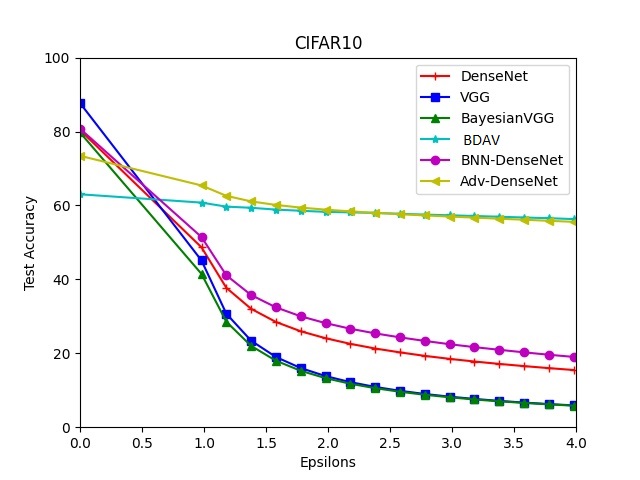}
        \includegraphics[width=0.329\linewidth]{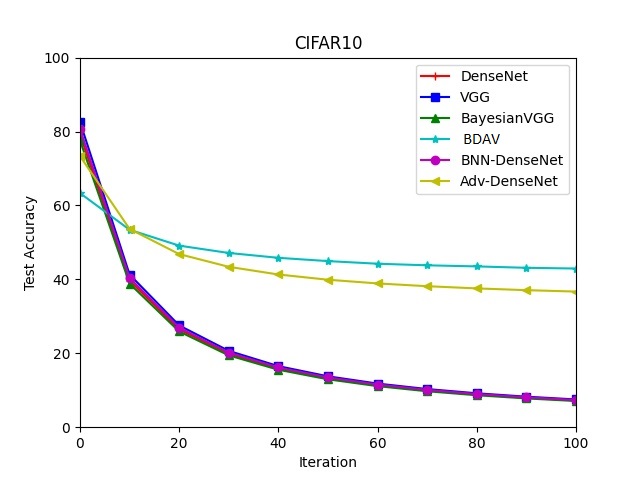}
    \end{subfigure}
    \caption{(Arranged from left-to-right) CIFAR10 trained models attacked with $l_{\infty}$-FGSM, 10 iterations $l_{\infty}$-PGD, 40 iterations $l_{\infty}$-PGD, 10 iterations $l_{2}$-PGD, 40 iterations $l_{2}$-PGD, and 
    Iteration vs. Test Accuracy for $l_{\infty}$-PGD}
    \label{fig:cifar10-attack}
\end{figure*}

\begin{figure*}[htb!]
    \centering
    \begin{subfigure}[b]{\textwidth}
        \centering
        \includegraphics[width=0.329\linewidth]{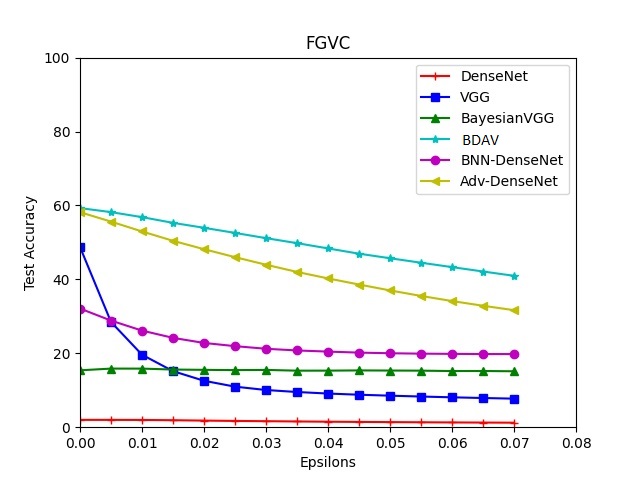}
        \includegraphics[width=0.329\linewidth]{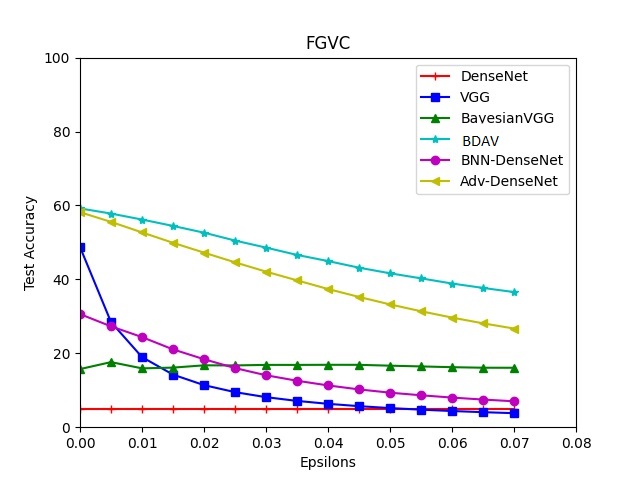}
        \includegraphics[width=0.329\linewidth]{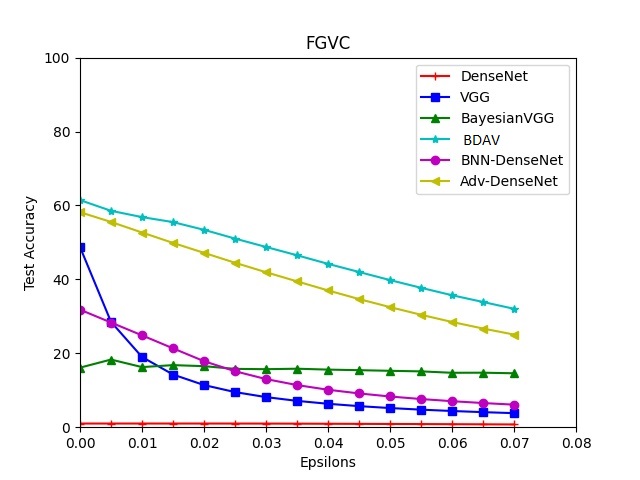}
    \end{subfigure}
    \begin{subfigure}[b]{\textwidth}
        \centering
        \includegraphics[width=0.329\linewidth]{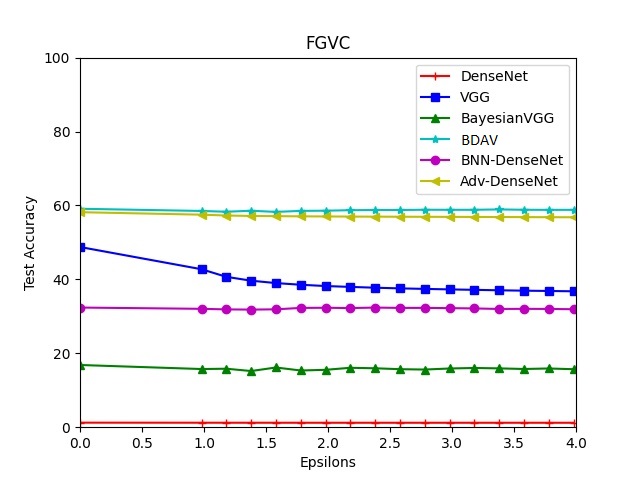}
        \includegraphics[width=0.329\linewidth]{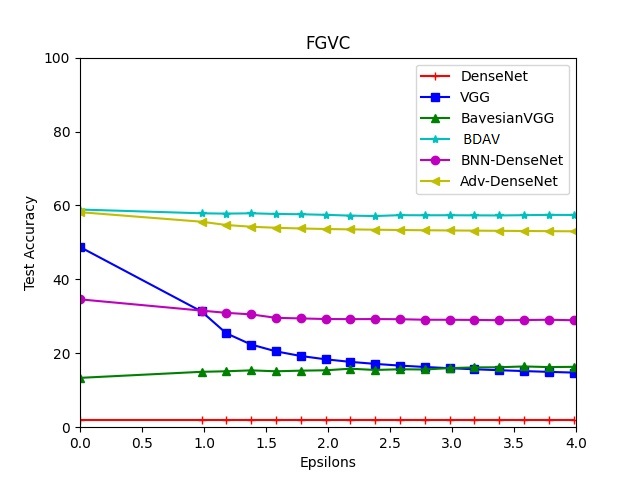}
        \includegraphics[width=0.329\linewidth]{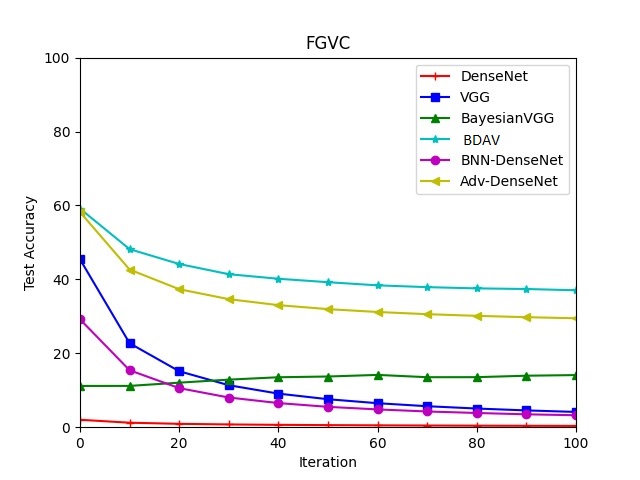}
    \end{subfigure}
    \caption{(Arranged from left-to-right) FGVC-Aircraft trained models attacked with $l_{\infty}$-FGSM, 10 iterations $l_{\infty}$-PGD, 40 iterations $l_{\infty}$-PGD, 10 iterations $l_{2}$-PGD, 40 iterations $l_{2}$-PGD, and 
    Iteration vs. Test Accuracy for $l_{\infty}$-PGD}
    \label{fig:FGVC-attack}
\end{figure*}

\begin{figure*}[htb!]
    \centering
    \begin{subfigure}[b]{\textwidth}
        \centering
        \includegraphics[width=0.329\linewidth]{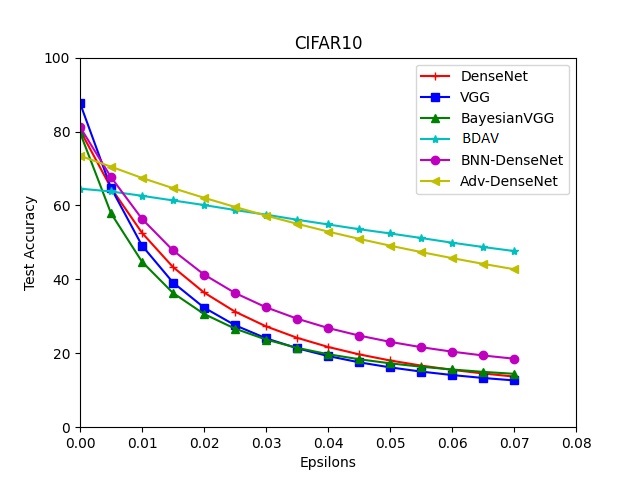}
        \includegraphics[width=0.329\linewidth]{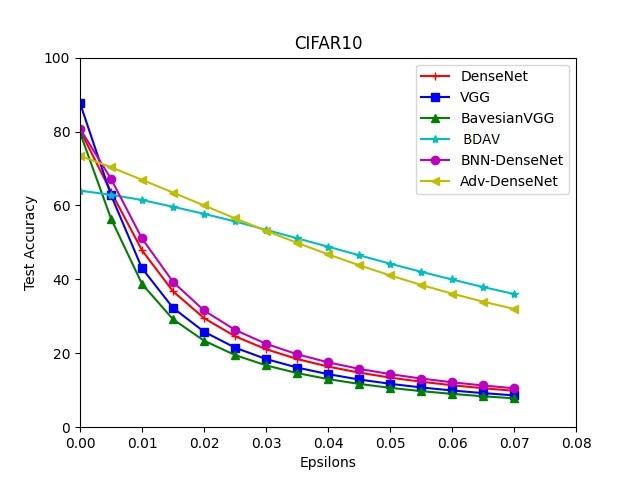}
    \end{subfigure}
    \begin{subfigure}[b]{\textwidth}
        \centering
        \includegraphics[width=0.329\linewidth]{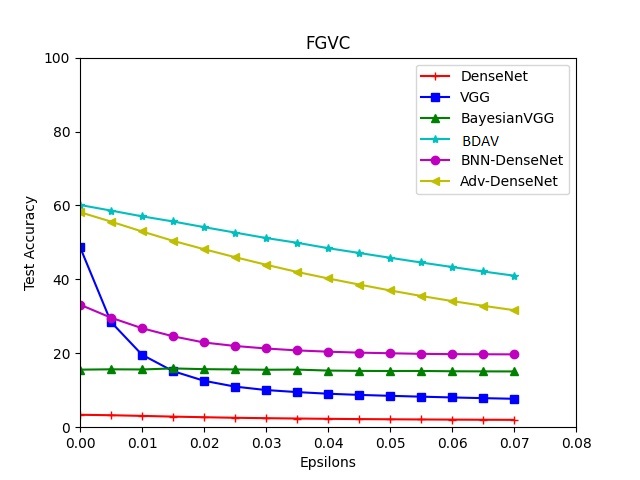}
        \includegraphics[width=0.329\linewidth]{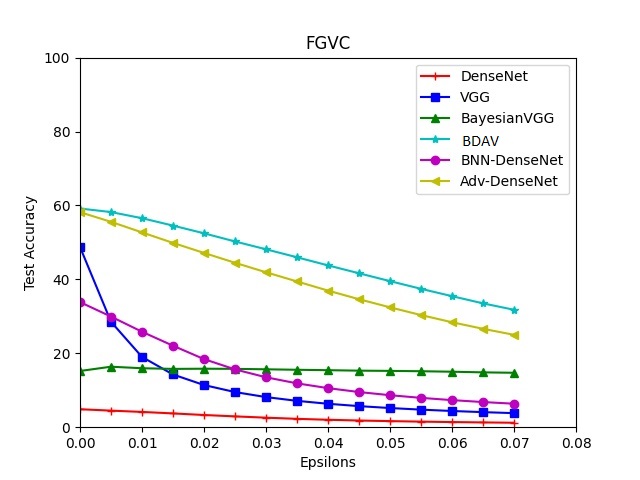}
    \end{subfigure}
    \caption{CIFAR-10 (top) and FGVC (bottom) trained models attacked with EOT $l_{\infty}$-FGSM and EOT $l_{\infty}$-PGD attacks with 30 ensemble size and 40 iterations}
    \label{fig:EOT}
\end{figure*}

\begin{figure*}[htb!]
    \centering
    \begin{subfigure}[b]{\textwidth}
        \centering
        \includegraphics[width=0.25\linewidth]{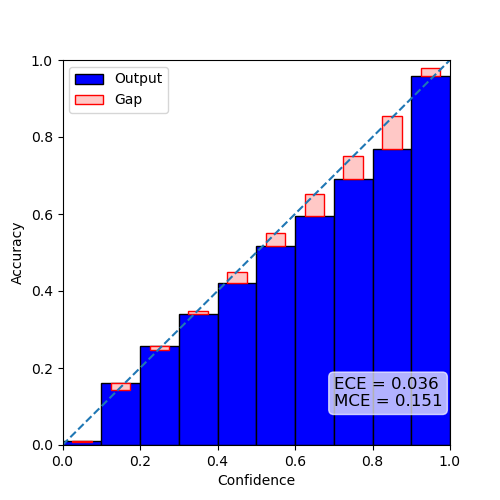}
        \includegraphics[width=0.25\linewidth]{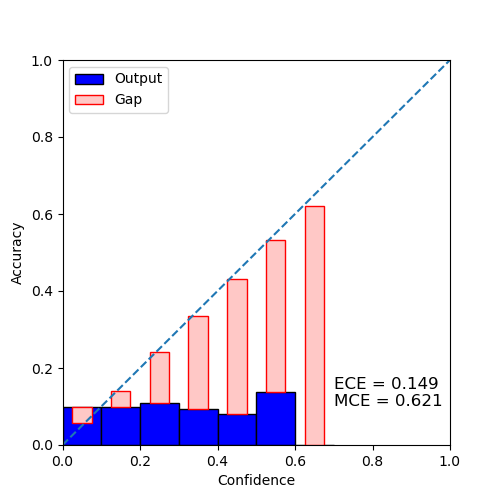}
        \includegraphics[width=0.25\linewidth]{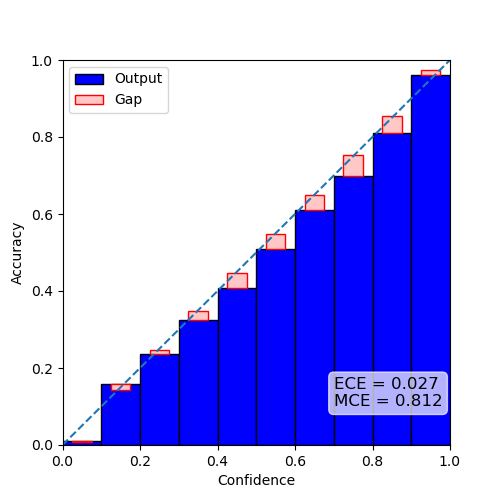}
    \vskip\baselineskip
        \includegraphics[width=0.25\linewidth]{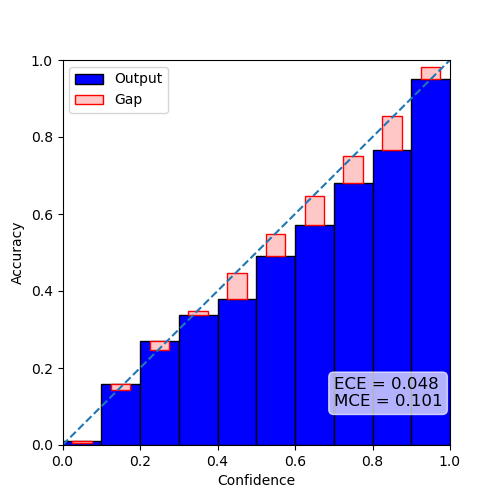}
        \includegraphics[width=0.25\linewidth]{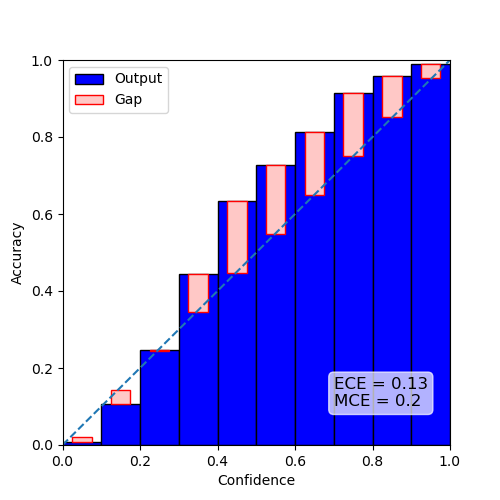}
        \includegraphics[width=0.25\linewidth]{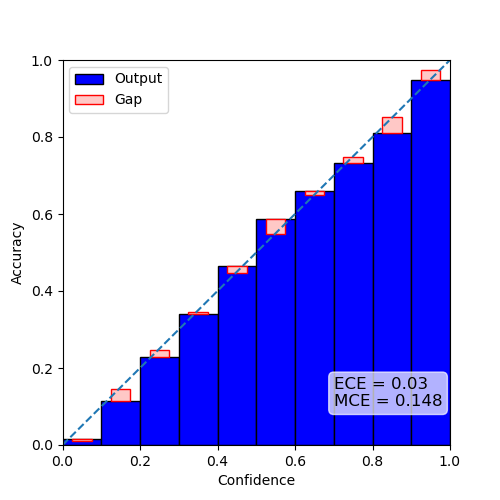}
    \subcaption{CIFAR-10}
    \end{subfigure}
    
    \vskip\baselineskip
     
     \begin{subfigure}[b]{\textwidth}
        \centering
        \includegraphics[width=0.25\linewidth]{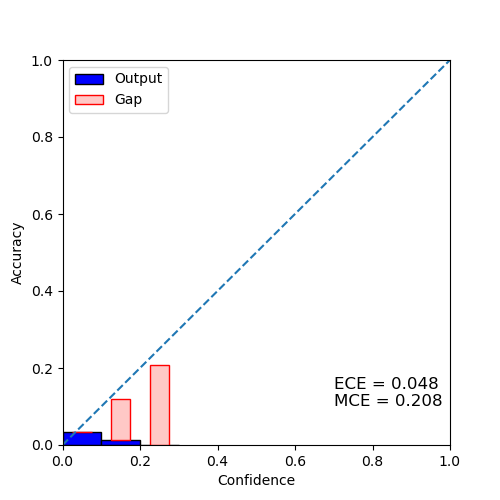}
        \includegraphics[width=0.25\linewidth]{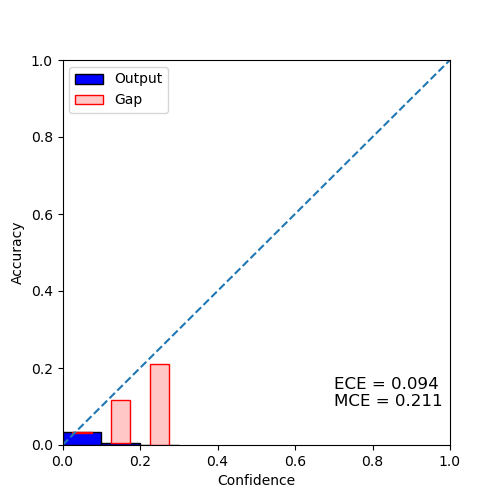}
        \includegraphics[width=0.25\linewidth]{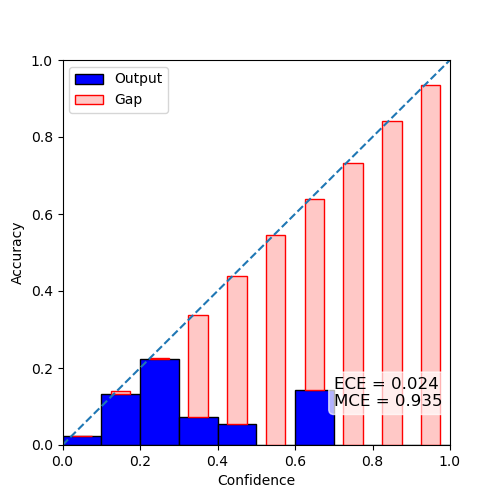}
    \vskip\baselineskip
        \includegraphics[width=0.25\linewidth]{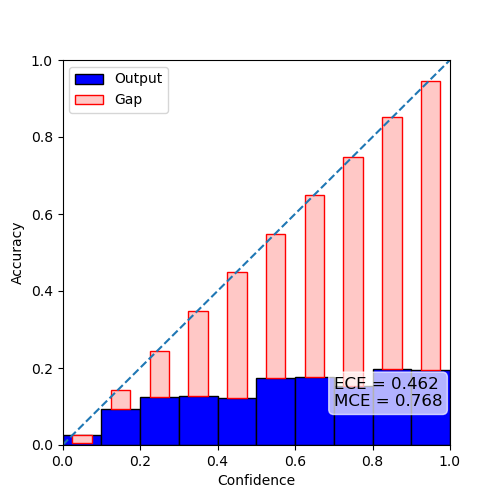}
        \includegraphics[width=0.25\linewidth]{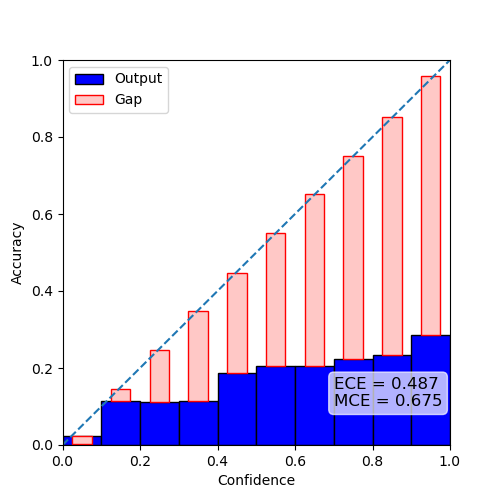}
        \includegraphics[width=0.25\linewidth]{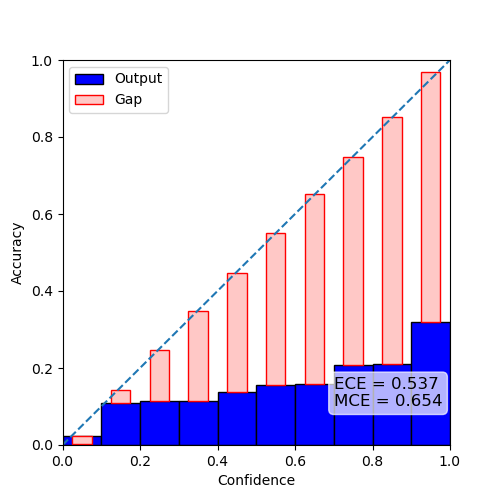}
    \subcaption{FGVC}

    \end{subfigure}
    
    \caption{(Arranged from left-to-right) Network calibration plots of DenseNet, VGG, BayesianVGG, BNN-DenseNet, BDAV, and Adv-DenseNet.}
    \label{fig:conf}
\end{figure*}

\section{Results}
In this section, we compare the performance of the baseline models 
to \textit{BNN-DenseNet} and \textit{BDAV}. 
Using CIFAR-10 and 
FGVC datasets, we train these 6 models and obtain the following test 
accuracies in Figure \ref{fig:acc}. 
For CIFAR-10, VGG achieves the highest test accuracy (87.700\%), and for FGVC, BDAV achieves the 
highest accuracy (58.566\%). 

We attack our trained models with strong state-of-the-art 
white-box adversarial attacks. See Figures \ref{fig:cifar10-attack} 
and \ref{fig:FGVC-attack}
for the performance of models trained on CIFAR-10 and FGVC attacked with $l_{\infty}$-FGSM, $l_{\infty}$-PGD, and
$l_{2}$-PGD, respectively. The figures are results of the following attacks:
$l_{\infty}$-FGSM, 10 iterations of $l_{\infty}$-PGD, 40 iterations of $l_{\infty}$-PGD, 10 iterations of $l_{2}$-PGD, 40 iterations of $l_{2}$-PGD, and 
Iteration (between 0-100) vs. Test Accuracy for $l_{\infty}$-PGD with 
$\epsilon=0.03$, respectively. 
We also attacked our models with EOT $l_{\infty}$-FGSM and EOT $l_{\infty}$-PGD attacks using 30 ensemble size and 40 iterations. 
See Figure \ref{fig:EOT} for plotted results of the models attacked with EOT attacks. 

Then, to further investigate the robustness of these models, we 
perform a network calibration analysis. 
We plot the accuracy's output and confidence (probabilities of each class)  into an identity function  
to evaluate each model's quality of the predictions.
Using 10 bins, we can see the difference between the 
model confidence and accuracy. See figure 
\ref{fig:conf} for network calibration plots. 

Lastly, we observe that the computational cost of the EOT attacks are 
high. The EOT $l_{\infty}$-PGD was especially expensive, taking about 
10 days to attack CIFAR-10 and 53 days to attack FGVC.

\section{Discussion}
The selection of a particular dataset can make a model seem robust. This
can be seen in Figures \ref{fig:cifar10-attack}, \ref{fig:FGVC-attack}
and \ref{fig:EOT}, where traditional NNs (VGG and DenseNet) perform 
reasonably on CIFAR-10 but underperform on FGVC. Therefore, to 
properly ascertain the robustness of a model, 
we find that training on complex data can be beneficial in extracting concepts.

Furthermore, we observe that BNN-DenseNet outperforms the traditional 
NNs and BayesianVGG in Figure \ref{fig:cifar10-attack}. This suggest
that BNNs are more robust to adversarial attacks than traditional NNs.
We also observe that BDAV outperforms Adv-DenseNet. 
What is most 
interesting here is that the two models intersect in performance at 
$\epsilon=0.03$. It is at this point that BDAV starts to outperform 
Adv-DenseNet.
However, in the case of FGVC, the performance of the BNNs are even 
more exaggerated. We see that in Figure \ref{fig:cifar10-attack} 
that for all attacks except for $l_2$-PGD with 10 iterations, a BNN 
model (BNN-DenseNet or BayesianVGG), outperforms all the traditional NNs,
significantly. 
Also, BDAV outperforms Adv-DenseNet in this scenario 
significantly as well. 
Therefore, the results suggest that since
BDAV decays linearly at a slow rate when attacked with strong epsilons,
it is much more able to withstand white-box adversarial attacks than other models. 

Additionally, to further evaluate the BNN models' robustness to adversarial attacks that incorporate randomization, we attack our trained models with EOT attacks. 
We find that for CIFAR-10, the difference between the naive attacks and EOT are relatively the same. This may be due to the dataset, not the model, because we observe a very different performance on models trained on FGVC. 
We see that our BNN models are more robust than traditional NNs in this attack scenario. 
Since BNN is a model that incorporates randomization, it is robust but should be susceptible to stochastic attacks in nature.  
However, we find that the opposite holds true as our BNN models
are robust to the EOT attacks. 
The reason for this behavior could be because BNNs operate under a mathematically justifiable principle. 
Using Bayesian Inference provides many benefits, such as the promotion of a stable stochastic model. 
Furthermore, it is important to consider 
the computational cost of EOT attacks, especially the 
EOT $l_{\infty}$-PGD attack. 

Lastly, robustness for this work is defined as a model's ability to be insusceptible or have a low susceptibility to strong adversarial attacks and make accurate predictions (i.e., confidence and accuracy should match). 
Therefore, we investigate our 6 models' calibration and find that BNN-DenseNet and BayesianVGG require less calibration than the other models. 
We also see that these BNN models make less overconfident predictions and are therefore less likely to be fooled by adversarial perturbations of images. Additionally, since there is no free lunch, the 
incorporation of adversarial training to a BNN also has disadvantages.
This can be seen in Figure \ref{fig:conf}, where BDAV is seen as a bit overconfident
and at times having a slightly higher ECE and/or MCE than Adv-DenseNet
while BNN-DenseNet is almost well-calibrated. 

\section{Conclusion}
Since the traditional framework of deep learning or neural network models are 
susceptible to adversarial attacks, we claim that this vulnerability is as a result of the shortcomings of the models. These shortcomings include - 
overconfidence, being data greedy, the need for regularization, and a huge potential to overfit. These disadvantages appear because deep learning models/traditional NNs do not measure for uncertainty well. Therefore we claim 
that incorporating an uncertainty measure will mitigate these shortcomings and 
as an added bonus, improve the robustness. 
This is achieved by 
fusing Bayesian Inference with a traditional NN (DenseNet121). 
The dense nature of DenseNet121, provides advantages that are 
amplified when fused with Bayesian Inference. 
Then to further improve BNN-DenseNet, we incorporate adversarial training to the model and 
construct \textit{BDAV}.

We compare these two models' performance to 4 baseline models - 2 traditional NN, a BNN, and adversarially trained NN.  
Then, we attack our models with 5 different 
strong state-of-the-art white-box attacks - $l_\infty$-FGSM, $l_\infty$-PGD, $l_2$-PGD, EOT $l_\infty$-FGSM, and EOT $l_\infty$-PGD.
To further evaluate robustness, we also performed a network calibration analysis of our models to ascertain which models were closer to 
being well-calibrated without incorporating any calibration techniques.
We find that the BNNs are more robust than the traditional NNs. 
Also, incorporating adversarial training to a BNN improves the robustness even further. 
However, while it improves the BNN's robustness to adversarial attacks, there appears to be a cost. 
This is seen in 
Figure \ref{fig:conf}, where BNN-DenseNet is closer to being 
well-calibrated than BDAV. We also observe that BNN-DenseNet has a lower ECE than BDAV. 


In the future, we will explore benefits of calibrating \textit{BDAV} to improve test accuracy and robustness. We will also 
investigate the effect of using different approximation functions,
like MCMC on adversarial robustness.

\section*{Acknowledgements}
This work was supported by the Robust and Secure Machine Learning (RSML) effort under Air Force Research Lab through the 
Autonomy Technology Research Center (ATRC) at Wright State University.

\bibliographystyle{IEEEtranN}
\bibliography{sample.bib}
\end{document}